%% file: main.tex
\documentclass[journal]{IEEEtran}
\usepackage{amsmath,amssymb}
\usepackage{booktabs}
\usepackage{cite}
\usepackage{graphicx}
\usepackage{multirow}
\usepackage{placeins}
\usepackage{float}
\usepackage{stfloats}
\usepackage{url}
\usepackage[table]{xcolor}


\newcommand{\method}{\textsc{TAPR}}

\ifCLASSINFOpdf
\else
\fi
\begin{document}
%
\title{Atomic Edit Factorization for Training-Free Composed Image Retrieval}
\author{Zihao Zhang, Dayan Wu, Xinze Liu, Hengjie Zhu, Ding Wang, Peng Fu, Zheng Lin,
Weiping Wang%
\thanks{(\emph{Corresponding author: Dayan Wu.})}%
\thanks{Z. Zhang, D. Wu, X. Liu, H. Zhu, P. Fu, Z. Lin, and W. Wang are with the
Institute of Information Engineering, Chinese Academy of Sciences, Beijing
100093, China (e-mail: zhangzihao@iie.ac.cn; wudayan@iie.ac.cn;
liuxinze@iie.ac.cn; zhuhengjie@iie.ac.cn; fupeng@iie.ac.cn; linzheng@iie.ac.cn;
wangweiping@iie.ac.cn).}%
\thanks{D. Wang is with the Department of Applied Mathematics and Statistics,
Johns Hopkins University, Baltimore, MD 21218, USA
(e-mail: dwang141@jh.edu).}}
\markboth{}{} 
%



\maketitle

\begin{abstract}
\input{abstract}
\end{abstract}


%
\IEEEpeerreviewmaketitle

\section{Introduction}
\input{introduction}

\section{Related Work}
\input{related_work}
\input{method}
\input{table_main_results}
\input{experiment}

\section{Conclusion}
\input{conclusion}

\FloatBarrier

\bibliographystyle{IEEEtran}
\bibliography{references}


%


\ifCLASSOPTIONcaptionsoff
  \newpage
\fi

\end{document}

%% file: abstract.tex
Training-free zero-shot composed image retrieval finds a target image in a
gallery from a reference image and a text edit without learning from
task-specific image triplets. Existing methods have achieved strong results by
using large language models to produce precise target descriptions. However,
matching each description only against a global image representation may mix
semantic cues and miss which attributes or relations
should change, remain, or disappear. We propose \method{}, which uses Atomic Edit
Factorization to convert an entangled edit request into four factors: Target,
Add, Preserve, and Remove. Frozen encoders represent the complete target state
with one description vector and the atomic edit constraints with factor-specific
probe vectors. Each candidate exposes a global visual vector and local visual vectors
as an evidence bank for these factors. Fine-grained multi-vector matching
connects each constraint to candidate evidence,
while factor-specific operators determine whether the matched evidence is
rewarded, checked against the reference, or penalized. Across CIRCO, CIRR, and
FashionIQ and three backbone scales, \method{}
improves CIRCO mAP@5 by up to 23.0\% over the strongest compared method and
gives the best CIRR and FashionIQ results in our comparison.

%% file: introduction.tex
\label{sec:introduction}

Zero-shot composed image retrieval (ZS-CIR) searches a gallery with a reference
image and a natural-language edit, without training on task-specific image
triplets. It provides a direct way to ask for an image like the reference but
with a requested change. FashionIQ, CIRR, and CIRCO test this task in
fine-grained fashion and open-domain settings~\cite{wu2021fashioniq,liu2021cirr,
baldrati2023circo}.

Recent training-free methods have achieved strong retrieval
without task-specific optimization by improving target descriptions
~\cite{karthik2024cirevl,yang2024ldre,tang2025osrcir,zhao2025dualcir} and
combining complementary visual, textual, or directional signals
~\cite{wen2024dqucir,li2025imagine,tursun2026pdv,psomas2025icir}. Further gains
come from local reranking and more structured retrieval evidence
~\cite{sun2023tfcir,sun2025cotmr,li2026stitch,jung2026soft,sun2026sdrcir}. A
correct result must satisfy the new request while keeping the relevant source
content. Since one edit can ask to add, keep, and remove different content at
the same time, the task requires more than overall image--text similarity.

\input{motivation_figure}

Existing training-free methods commonly describe the desired target and
retrieve images close to that description~\cite{karthik2024cirevl,
tang2025osrcir}. Later methods add positive and negative descriptions,
directional scores, reference-semantic debiasing, or local set
matching~\cite{zhao2025dualcir,sun2025cotmr,tursun2026pdv,li2026stitch,
jung2026soft,sun2026sdrcir}. These designs improve retrieval, but they still
treat a composed request as an entangled specification rather than separating
its distinct edit functions. Figure~\ref{fig:motivation} illustrates
this limitation using CIRR validation query 37730. The edit asks to make the
dog older, place two birds next to it, and turn the scene into a painting. The
left branch shows CoTMR's global description matching~\cite{sun2025cotmr}.
CoTMR forms a precise target description that includes all three changes and
compares it with one global visual vector for each candidate. However, this
single comparison lets the dominant dog content outweigh the missing birds and
painting style, producing the wrong result. The failure exposes the central
limitation: even a precise description can lose a small but decisive detail
when distinct edit functions are compressed into one global match. It also
cannot distinguish evidence for content that should be added, continuity that
should be preserved from the reference, and a source state that should be
removed.

To address these limitations, we propose \method{}, which connects composed
edits to candidate evidence through Atomic Edit Factorization. As shown in the
right branch of Figure~\ref{fig:motivation}, a frozen factorizer separates the
request into Target, Add, Preserve, and Remove factors. Target anchors the
complete edited scene, including the older dog and painting style; Add isolates
the two requested birds; Preserve keeps the seated dog; and Remove identifies
realistic photographic cues as counter-evidence. Compound constraints are split
into atoms and encoded as factor-specific probe vectors, while Target is encoded
as one target description vector. Each candidate forms an evidence bank
containing one global visual vector and local visual vectors. Fine-grained
multi-vector matching checks small edit atoms against local evidence, and
factor-specific operators reward Add evidence, compare Preserve evidence with
the reference, and penalize Remove evidence. This design prevents distinct edit
requirements from collapsing into one representation and provides the local
evidence needed to verify them independently. In the example, TAPR prevents the
dog match from masking the missing constraints and retrieves the ground-truth
target at rank one.

Our contributions are threefold:
\begin{itemize}
  \item We propose Atomic Edit Factorization, which organizes a composed edit
  request into Target, Add, Preserve, and Remove factors at the granularity of
  individual visual constraints. It avoids semantic entanglement among distinct edit
  requirements during candidate matching and gallery ranking.

  \item We build \method{}, a training-free framework that connects the
  factorized edit specification to candidate evidence. It matches a target
  description vector and factor-specific probe vectors against global and local
  visual vectors. This resolves the mismatch between fine-grained edit
  constraints and holistic global matching.

  \item Across CIRCO, CIRR, FashionIQ, and three backbone scales, \method{}
  improves CIRCO mAP@5 by up to 23.0\% over the strongest baseline and achieves
  the best CIRR and FashionIQ results in our comparison. Matched controls
  further show that factor-wise matching improves retrieval with either global
  evidence alone or combined global and local evidence.
\end{itemize}

%% file: motivation_figure.tex
\begin{figure}[!t]
  \centering
  \includegraphics[width=\columnwidth]{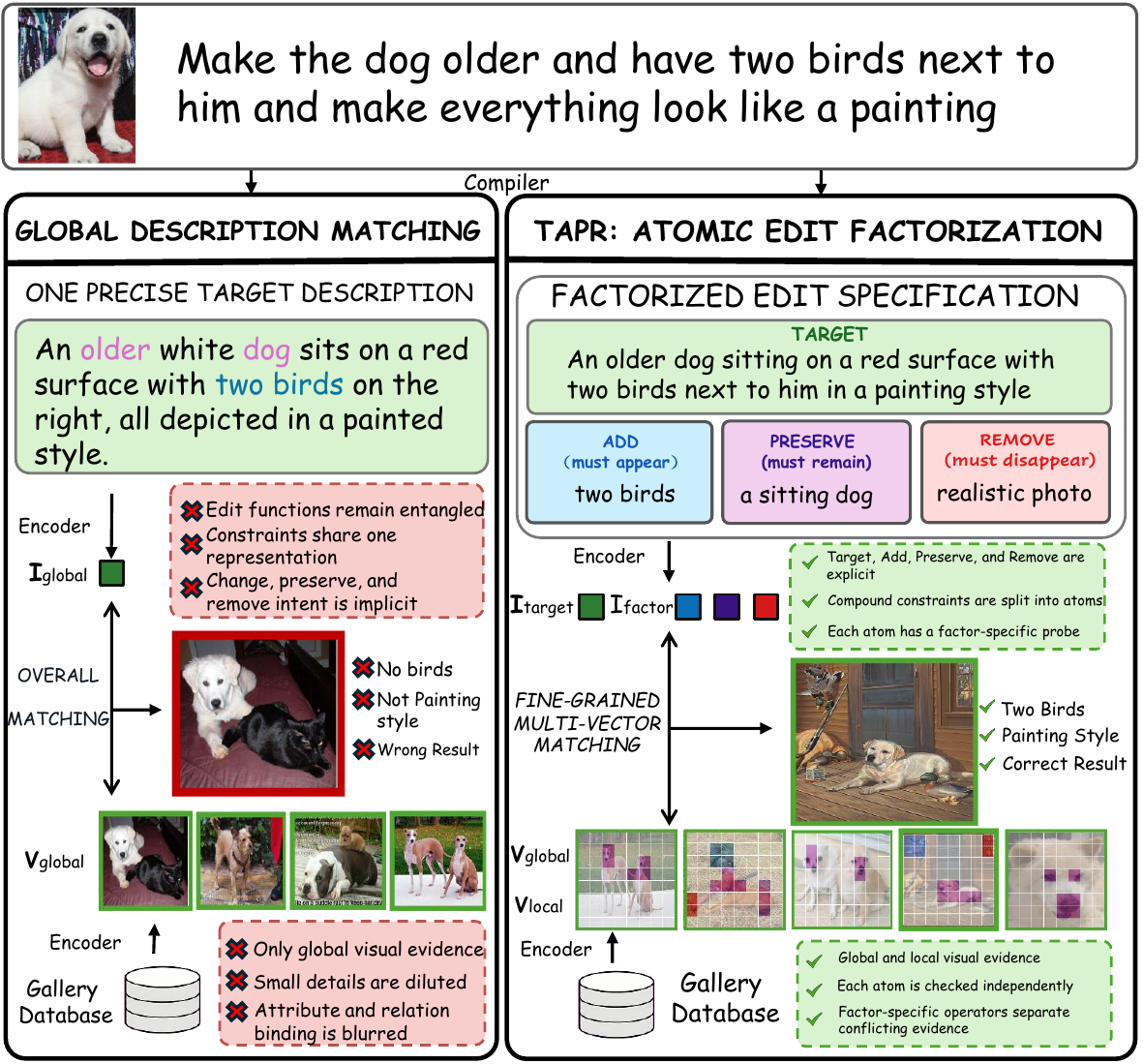}
  \caption{Motivation on CIRR validation query 37730. Left: CoTMR's global
  description matching~\cite{sun2025cotmr} encodes one precise target
  description and compares it with one global visual vector per candidate, but
  its retrieval misses the requested birds and painting style. Right: \method{}
  factorizes the request into Target, Add, Preserve, and Remove factors, splits
  their constraints into atoms, and checks factor-specific probes against
  global and local candidate evidence to retrieve the ground-truth target at
  rank one.}
  \label{fig:motivation}
\end{figure}

%% file: related_work.tex
\label{sec:related_work}

\subsection{Learned CIR and Zero-Shot Transfer}

Composed image retrieval (CIR) retrieves a target image from a reference image
and a text modification. Supervised methods learn this composition from
annotated reference--text--target triplets. Early work maps the two query
modalities into one embedding or a shared compositional space
\cite{vo2019tirg,chen2020jvsm,anwaar2021composeae,kim2021dcnet}, while later
methods introduce region--word matching, CLIP-based residual fusion,
candidate-aware reranking, concept supervision, or multimodal language models
\cite{hosseinzadeh2020local,baldrati2022clip4cir,liu2024candidate,
xing2025contextcir,huynh2025collm}.

Zero-shot CIR removes annotated triplets from the target domain, but does not
necessarily remove task-specific learning. Representative methods learn an
image-to-word mapping, textual inversion, a language-only projection,
knowledge-enhanced fusion, diffusion-based composition, text-anchored tuning,
instruction-aware distillation, or self-guided semantic inspection
\cite{saito2023pic2word,baldrati2023circo,gu2024lincir,lin2024fti4cir,
suo2024keds,gu2024compodiff,jang2024slerptat,zhong2025distillcir,
zhang2026diffcomp}. These methods improve transfer to unseen compositions, but
remain outside our setting because they optimize a CIR-specific mapping,
encoder, or composition module. We use \,\emph{training-free}\, in the stricter
sense that retrieval is built from frozen pretrained models without such
optimization.

\subsection{Training-Free Zero-Shot CIR}
\label{sec:training_free_zscir}

Training-free ZS-CIR constructs the query and its retrieval scores at
inference. CIReVL captions the reference image and asks a language model to
rewrite the caption according to the modification
\cite{karthik2024cirevl}. TFCIR makes this query explicit and adds local-concept
reranking \cite{sun2023tfcir}. LDRE generates diverse edited captions and
combines them according to semantic relevance \cite{yang2024ldre}, whereas
OSrCIR directly reasons over the reference image and modification to produce a
target description in one stage \cite{tang2025osrcir}. These methods mainly
improve how target-directed descriptions are produced or combined.

Other methods retain more than one retrieval signal. DQU-CIR applies two
training-free raw-data transformations to form a textual query and a visually
modified query before combining their features \cite{wen2024dqucir}. DualCIR
uses dual-directional descriptions \cite{zhao2025dualcir}; Imagine and Seek
retrieves with a generated proxy image \cite{li2025imagine}; and PDV transfers
a prompt direction to text and image embeddings \cite{tursun2026pdv}. BASIC
separately scores reference-to-candidate and text-to-candidate similarity, then
uses late fusion to reward candidates that satisfy both signals
\cite{psomas2025icir}. These designs preserve complementary query signals, but
do not factorize the requested change into distinct semantic functions with
their own matching behavior.

Recent methods further structure the retrieval evidence. CoTMR reasons with
global and object-scale descriptions of existent and nonexistent content
\cite{sun2025cotmr}. STiTch combines semantic transition with set-to-set
transportation \cite{li2026stitch}. SoFT extracts prescriptive and proscriptive
constraints and uses them as reward and penalty filters for a base retriever
\cite{jung2026soft}. SDR-CIR guides reference understanding with Selective CoT,
anchors the composed query with reference-image features, and penalizes the
estimated reference-induced bias in the final score \cite{sun2026sdrcir}.
These works establish the value of direction, scale, signed constraints,
reference-aware correction, and set matching. However, these components address
different parts of the query without a shared factorization that connects each
constraint to its matching behavior. The remaining issue is to
factorize an entangled target specification into Target, Add, Preserve, and
Remove functions, atomize their edit constraints, and give each factor the
matching rule required by its meaning. This alignment keeps each factor tied
to its required evidence.

\begin{figure*}[!t]
  \centering
  \includegraphics[width=\textwidth]{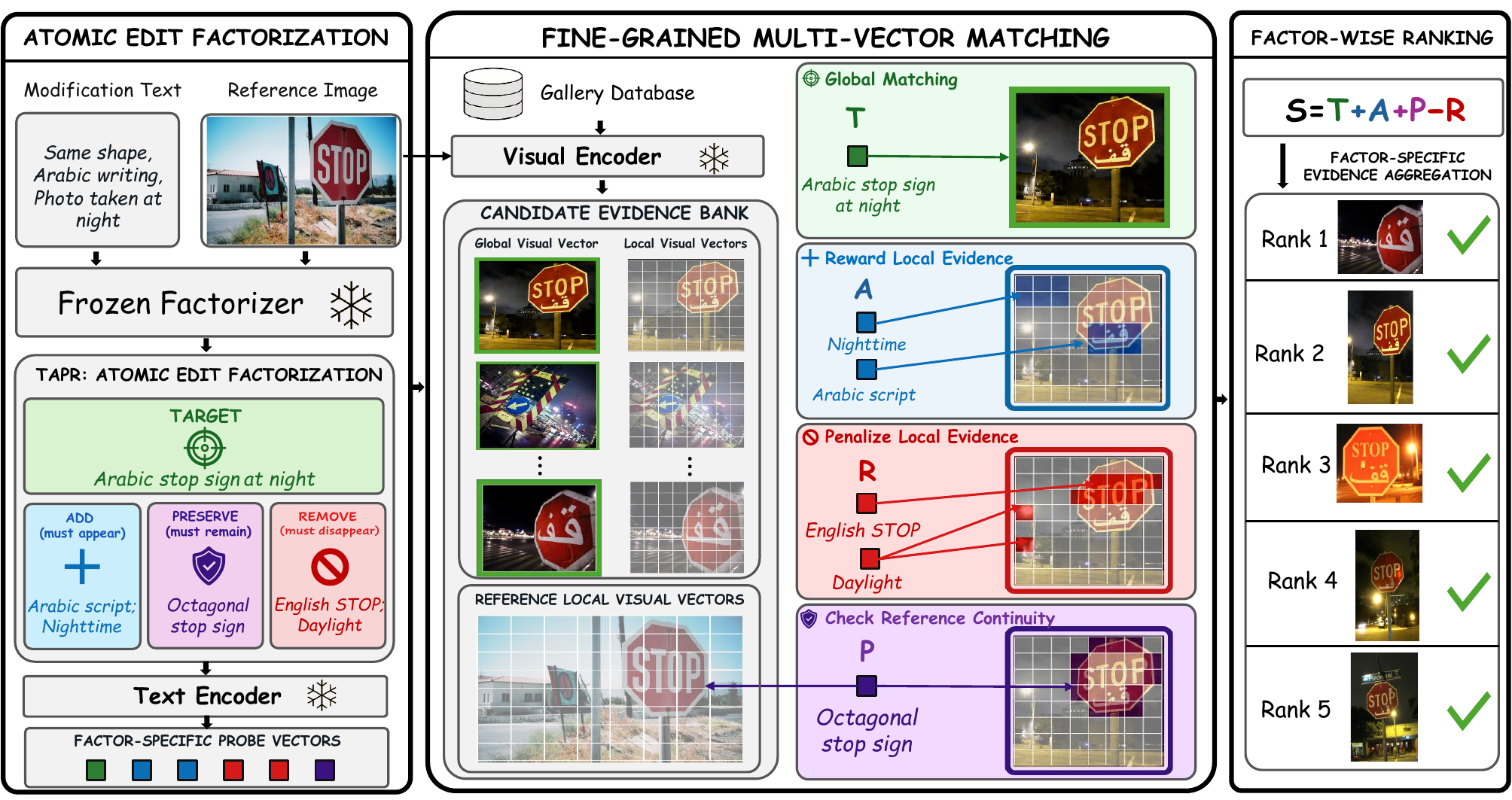}
  \caption{The \method{} pipeline on CIRCO validation query 45. Atomic Edit
  Factorization maps the reference image and modification to a factorized edit
  specification: Target describes an Arabic stop sign at night, Add specifies
  Arabic script and nighttime, Preserve keeps the octagonal stop-sign structure,
  and Remove suppresses English ``STOP'' and daylight. Frozen encoders produce
  one target description vector, factor-specific probe vectors, a candidate
  evidence bank of global and local visual vectors, and reference local visual
  vectors. Fine-grained multi-vector matching uses a different operator for each
  factor: Target performs global matching, Add rewards local evidence, Preserve
  checks reference continuity, and Remove penalizes local evidence. Factor-wise
  aggregation ranks the gallery; all five displayed results are ground-truth
  positives.}
  \label{fig:method_pipeline}
\end{figure*}

\subsection{Fine-Grained and Structured Evidence Matching}

Fine-grained candidate evidence is not new by itself. Region--word interaction
was introduced in supervised CIR \cite{hosseinzadeh2020local}, and
candidate-aware reranking, local-concept reranking, object-scale scoring, and
set transportation later brought related interactions to learned and
training-free CIR \cite{liu2024candidate,sun2023tfcir,sun2025cotmr,
li2026stitch}. Outside CIR, ColBERT and FILIP are established examples of
token-level and token--patch late interaction
\cite{khattab2020colbert,yao2021filip}. These precedents show why local or
set-valued features can expose small visual details. They do not, however,
specify how an entangled edit should be factorized into independently
verifiable functions.

Our distinction is Atomic Edit Factorization followed by fine-grained
multi-vector matching with factor-specific operators. SoFT and SDR-CIR are the
closest works along complementary axes. SoFT
separates must-have and must-avoid constraints but applies global reward and
penalty filters
\cite{jung2026soft}; SDR-CIR makes ranking reference-aware but estimates one
reference-induced bias term around a composed target description
\cite{sun2026sdrcir}. Neither treats Preserve as a separate factor with its own
reference-dependent local operator. In \method{}, Target, Add, Preserve,
and Remove are separated, and the edit constraints are represented as edit
atoms. Add rewards local support, Preserve checks local continuity with the
reference, and Remove penalizes local support, all over the same candidate
evidence bank. Fine-grained multi-vector matching over this bank provides the
local evidence needed to verify the factorized atoms.

%% file: method.tex
\section{Method}
\label{sec:method}

\subsection{Problem Formulation}
\label{sec:problem_formulation}

Given a reference image $I_r$ and a natural-language modification $m$, a
composed query is $q=(I_r,m)$. For a gallery
$\mathcal G=\{I_1,\ldots,I_N\}$, the system assigns each candidate a score
$S(q,I)$ and returns the descending ranking
\begin{equation}
 \pi_q=\operatorname*{argsort}_{I\in\mathcal G} S(q,I).
 \label{eq:cir_task}
\end{equation}
The target and its relevance labels are unavailable at inference. In the
training-free zero-shot setting, no task-specific parameters are learned. A
valid result must satisfy the requested change while preserving the relevant
source content.

\subsection{TAPR Overview}
\label{sec:method_overview}

\method{} first factorizes a composed edit and then matches each factor to
candidate evidence. The Target factor specifies the complete result, Add
evidence should increase a candidate's score, Preserve evidence should be checked
relative to the reference, and Remove evidence should decrease the score.
Composite Add, Preserve, and Remove requirements are further decomposed into
atomic visual constraints.

Figure~\ref{fig:method_pipeline} illustrates the three-stage flow on CIRCO
validation query 45. First, Atomic Edit Factorization and a frozen text encoder
map $(I_r,m)$ to a target description vector and Add, Preserve, and Remove
factor-specific probe vectors. Next, a frozen image encoder builds a candidate
evidence bank containing one global visual vector and local visual vectors,
together with local visual vectors for the reference. Fine-grained multi-vector
matching then applies the factor-specific operators shown in the figure: Target
performs global matching, Add rewards evidence of Arabic script and nighttime,
Preserve checks the octagonal shape against the reference, and Remove penalizes
English ``STOP'' and daylight. Factor-wise ranking aggregates these signals once
over the full gallery; all five displayed results are ground-truth positives.
The multi-vector evidence bank exposes atom-level candidate evidence, while the
factor-specific operators determine how each match contributes to the score.

\subsection{Atomic Edit Factorization}
\label{sec:query_compilation}

A frozen factorizer receives only $I_r$ and $m$ and directly produces four
semantic factors:
\begin{equation}
 \mathcal B=(T,\mathcal A,\mathcal P,\mathcal R),
 \label{eq:role_bundle}
\end{equation}
where $T$, $\mathcal A$, $\mathcal P$, and $\mathcal R$ denote the Target,
Add, Preserve, and Remove factors, respectively.
Target describes the complete desired image; Add lists content that must
appear; Preserve lists source identity or context that must remain; and Remove
lists source states that must disappear or be replaced. Each element of
$\mathcal A$, $\mathcal P$, and $\mathcal R$ represents one atomic visual
constraint. The frozen text tower encodes Target as the target
description vector $t(T)$ and encodes the edit atoms as factor-specific probe
sets $U_A=t(\mathcal A)$, $U_P=t(\mathcal P)$, and
$U_R=t(\mathcal R)$. Each probe states what evidence to test, while its factor
determines how a match contributes to the final score.
The factorizer never accesses a target image, target identifier, candidate, or
relevance label.

\subsection{Fine-Grained Multi-Vector Matching}
\label{sec:role_retrieval}

Edit atoms can depend on small visual details that a global visual vector may
compress away. We therefore use one frozen vision--language encoder
\cite{radford2021clip}. Its text tower maps a phrase $x$ to a unit vector
$t(x)\in\mathbb R^d$, and its image tower stores each candidate as one global
visual vector plus $M$ local visual vectors. Together they form the candidate
evidence bank
\begin{equation}
\begin{aligned}
 \mathcal E(I)&=\bigl(v_0(I),V_\ell(I)\bigr),\\
 V_\ell(I)&=\{v_j(I)\}_{j=1}^{M},\qquad
 \|v_0(I)\|_2=\|v_j(I)\|_2=1,
\end{aligned}
 \label{eq:image_vectors}
\end{equation}
where $v_0$ is the global visual vector and $V_\ell$ is the set of local visual
vectors.
Fine-grained multi-vector matching compares the factor-specific probes with
both global and local vectors in this bank. The bank changes the granularity of
available evidence; the factor-specific operators below determine its meaning.

The local operators share one smooth late-interaction primitive, following the
general idea of fine-grained late interaction
\cite{khattab2020colbert,yao2021filip}. For unit factor-specific probe vectors
$U=\{u_i\}_{i=1}^{n}$ and a bank of candidate local visual vectors
$V=\{v_j\}_{j=1}^{M}$, we define
\begin{align}
 \pi_{ij}^{(\tau)}&=\frac{\exp(u_i^\top v_j/\tau)}
 {\sum_{k=1}^{M}\exp(u_i^\top v_k/\tau)},\\
 \mathcal L_\tau(U,V)&=\frac{1}{n}\sum_i\sum_j
 \pi_{ij}^{(\tau)}u_i^\top v_j .
 \label{eq:soft_local_pooling}
\end{align}
The soft weighting lets each probe gather local support without allowing one
accidental patch match to dominate.

Fine-grained multi-vector matching supplies the evidence, and factor-wise
operator dispatch determines its meaning. The four factors read the same
candidate evidence with different aggregation signs and reference
dependencies. Target matches the target description vector to the candidate
global visual vector:
\begin{equation}
 \Phi_T(I)=t(T)^\top v_0(I).
 \label{eq:global_operator}
\end{equation}
The Add and Remove operators are evaluated only for nonempty probe sets; an
empty set deactivates the corresponding factor. Add rewards local evidence for
the content that must appear:
\begin{equation}
 \Phi_A(I)=\mathcal L_\tau(U_A,V_\ell(I)).
 \label{eq:desired_operator}
\end{equation}
Preserve is reference-dependent. We write $C(I_r,I)$ for the match from the
reference local visual vectors $V_\ell(I_r)$ to the candidate local visual
vectors $V_\ell(I)$ under $\mathcal L_\tau$. Preserve combines
this visual continuity with textual support when Preserve probes are available
and otherwise uses continuity alone:
\begin{equation}
 \Phi_P(I_r,I)=
 \begin{cases}
  \tfrac12\mathcal L_\tau(U_P,V_\ell(I))+\tfrac12 C(I_r,I),
    & |U_P|>0,\\
  C(I_r,I), & |U_P|=0.
 \end{cases}
 \label{eq:preserve_operator}
\end{equation}
Remove measures local evidence for source states that must disappear:
\begin{equation}
 \Phi_R(I)=\mathcal L_\tau(U_R,V_\ell(I)).
 \label{eq:forbidden_operator}
\end{equation}
These operators produce the factor-wise score
\begin{equation}
\begin{aligned}
 S_{\mathrm{FW}}(q,I)
 &=\theta_T\Phi_T(I)+\theta_A\Phi_A(I)\\[-2pt]
 &\quad+\theta_P\Phi_P(I_r,I)-\theta_R\Phi_R(I).
\end{aligned}
 \label{eq:role_aware_score}
\end{equation}
Here $\theta_\rho\geq0$ for $\rho\in\{T,A,P,R\}$. Inactive Add and Remove
factors are omitted, and active weights are renormalized. Sorting
Eq.~\eqref{eq:role_aware_score} once over $\mathcal G$ produces the final ranking;
there are no parallel routes or rank fusion. The factor-to-operator mapping
defines how matches affect ranking, while fine-grained multi-vector matching
provides the local visual evidence needed to verify individual edit atoms. The
same factor-wise formulation also remains valid with a single global visual
vector.

%% file: table_main_results.tex
\begin{table*}[!t]
  \centering
  \caption{Direct single-pass training-free ZS-CIR on the CIRCO and CIRR test
  sets, both evaluated online with hidden test annotations. Bold and underlined
  values are the best and second best among the listed rows within each
  backbone block.}
  \label{tab:main_results}
  \setlength{\tabcolsep}{2.8pt}
  \renewcommand{\arraystretch}{1.13}
  \resizebox{\textwidth}{!}{%
  \begin{tabular}{@{}ll cccc ccc ccc@{}}
    \toprule
    \multirow{3}{*}{\textbf{Backbone}} &
    \multirow{3}{*}{\textbf{Method}} &
    \multicolumn{4}{c}{\textbf{CIRCO}} &
    \multicolumn{6}{c}{\textbf{CIRR}} \\
    \cmidrule(lr){3-6}\cmidrule(lr){7-12}
    & & \multicolumn{4}{c}{\textbf{mAP@$k$}} &
    \multicolumn{3}{c}{\textbf{Recall@$k$}} &
    \multicolumn{3}{c}{\textbf{Recall$_{sub}$@$k$}} \\
    \cmidrule(lr){3-6}\cmidrule(lr){7-9}\cmidrule(lr){10-12}
    & & 5 & 10 & 25 & 50 & 1 & 5 & 10 & 1 & 2 & 3 \\
    \midrule

    \multirow{8}{*}{ViT-B}
      & CIReVL~\cite{karthik2024cirevl} (ICLR'24)
      & 14.94 & 15.42 & 17.00 & 17.82 & 23.94 & 52.51 & 66.00 & 60.17 & 80.05 & 90.19 \\
      & OSrCIR~\cite{tang2025osrcir} (CVPR'25)
      & 18.04 & 19.17 & 20.94 & 21.85 & 25.42 & 54.54 & 68.19 & 62.31 & 80.86 & 91.13 \\
      & CoTMR~\cite{sun2025cotmr} (ICCV'25)
      & 22.23 & 22.78 & 24.68 & 25.74 & 31.50 & 60.80 & 73.04 & 66.61 & 84.50 & 92.55 \\
      & CIReVL + SoFT~\cite{jung2026soft} (AAAIW'26)
      & 19.21 & 20.04 & 21.86 & 22.75 & 32.94 & 62.92 & 74.17 & \underline{70.31} & 86.58 & 93.78 \\
      & LDRE + PDV-F~\cite{yang2024ldre,tursun2026pdv} (WACV'26)
      & 17.80 & 18.78 & 20.61 & 21.56 & 29.30 & 60.39 & 72.51 & 63.06 & 82.36 & 91.54 \\
      & SDR-CIR~\cite{sun2026sdrcir} (WWW'26)
      & \underline{23.78} & \underline{24.43} & \underline{26.58} & \underline{27.50}
      & \underline{34.48} & \underline{65.74} & \underline{76.87}
      & 69.90 & \underline{87.04} & \underline{94.48} \\
      & STiTch~\cite{li2026stitch} (CVPR'26)
      & 20.26 & 21.01 & 23.01 & 24.04 & 25.83 & 55.25 & 70.20 & 65.64 & 83.60 & 92.80 \\
      \rowcolor{gray!15}
      & \textbf{TAPR (Ours)}
      & \textbf{29.25} & \textbf{29.78} & \textbf{31.96} & \textbf{32.98}
      & \textbf{36.70} & \textbf{68.53} & \textbf{80.15}
      & \textbf{72.19} & \textbf{88.34} & \textbf{94.80} \\
    \midrule

    \multirow{8}{*}{ViT-L}
      & CIReVL~\cite{karthik2024cirevl} (ICLR'24)
      & 18.57 & 19.01 & 20.89 & 21.80 & 24.55 & 52.31 & 64.92 & 59.54 & 79.88 & 89.69 \\
      & OSrCIR~\cite{tang2025osrcir} (CVPR'25)
      & 23.87 & 25.33 & 27.84 & 28.97 & 29.45 & 57.68 & 69.86 & 62.12 & 81.92 & 91.10 \\
      & CoTMR~\cite{sun2025cotmr} (ICCV'25)
      & 27.61 & 28.22 & 30.61 & 31.70 & 35.02 & 64.75 & 76.18 & 69.39 & 85.75 & 93.33 \\
      & CIReVL + SoFT~\cite{jung2026soft} (AAAIW'26)
      & 23.90 & 24.72 & 26.94 & 27.93 & 35.54 & 65.25 & 76.41 & 71.59 & 87.64 & 94.15 \\
      & LDRE + PDV-F~\cite{yang2024ldre,tursun2026pdv} (WACV'26)
      & 25.23 & 26.52 & 28.94 & 29.95 & 30.16 & 59.98 & 71.90 & 63.66 & 82.87 & 91.57 \\
      & SDR-CIR~\cite{sun2026sdrcir} (WWW'26)
      & \underline{30.91} & \underline{31.50} & \underline{34.03} & \underline{35.08}
      & \underline{37.61} & \underline{67.71} & \underline{79.13}
      & \underline{71.90} & \underline{88.39} & \underline{94.63} \\
      & STiTch~\cite{li2026stitch} (CVPR'26)
      & 25.55 & 26.27 & 28.81 & 29.99 & 28.87 & 57.97 & 69.90 & 65.22 & 84.10 & 92.37 \\
      \rowcolor{gray!15}
      & \textbf{TAPR (Ours)}
      & \textbf{37.83} & \textbf{38.46} & \textbf{41.44} & \textbf{42.50}
      & \textbf{40.17} & \textbf{71.23} & \textbf{81.35}
      & \textbf{76.29} & \textbf{90.36} & \textbf{95.78} \\
    \midrule

    \multirow{7}{*}{ViT-G}
      & CIReVL~\cite{karthik2024cirevl} (ICLR'24)
      & 26.77 & 27.59 & 29.96 & 31.03 & 34.65 & 64.29 & 75.06 & 67.95 & 84.87 & 93.21 \\
      & OSrCIR~\cite{tang2025osrcir} (CVPR'25)
      & 30.47 & 31.14 & 35.03 & 36.59 & 37.26 & 67.25 & 77.33 & 69.22 & 85.28 & 93.55 \\
      & CoTMR~\cite{sun2025cotmr} (ICCV'25)
      & 32.23 & 32.72 & 35.60 & 36.83 & 36.36 & 67.52 & 77.82 & 71.19 & 86.34 & 93.87 \\
      & LDRE + PDV-F~\cite{yang2024ldre,tursun2026pdv} (WACV'26)
      & \underline{34.88} & \underline{36.41} & \underline{39.12} & \underline{40.23} & \underline{42.51} & \underline{72.22} & \underline{81.71} & 72.39 & 88.34 & 94.80 \\
      & SDR-CIR~\cite{sun2026sdrcir} (WWW'26)
      & 33.05 & 34.50 & 37.21 & 38.42 & 40.17 & 69.76 & 79.88 & 73.30 & 88.89 & 94.99 \\
      & STiTch~\cite{li2026stitch} (CVPR'26)
      & 34.40 & 35.56 & 38.07 & 40.02 & 39.23 & 69.95 & 79.56 & \underline{73.56} & \underline{89.50} & \underline{95.86} \\
      \rowcolor{gray!15}
      & \textbf{TAPR (Ours)}
      & \textbf{41.76} & \textbf{43.27} & \textbf{46.27} & \textbf{47.29}
      & \textbf{42.70} & \textbf{73.04} & \textbf{82.84}
      & \textbf{76.92} & \textbf{90.58} & \textbf{96.07} \\
    \bottomrule
  \end{tabular}%
  }
\end{table*}

%% file: experiment.tex
\section{Experiments}
\label{sec:experiments}

\subsection{Experimental Setup}
\label{sec:experimental_setup}

We evaluate on FashionIQ~\cite{wu2021fashioniq},
CIRR~\cite{liu2021cirr}, and CIRCO~\cite{baldrati2023circo}. CIRCO reports
mAP@$k$; CIRR reports full-gallery Recall@$k$ and Recall$_{sub}$@$k$ within its
official groups; FashionIQ reports Recall@10/50 for Shirt, Dress, and Toptee.
Tables~\ref{tab:main_results} and~\ref{tab:fashioniq_results} summarize the
corresponding public comparisons.

The experiments are organized to test the paper's central claim that Atomic
Edit Factorization gives consistent retrieval gains, while fine-grained
multi-vector matching improves the evidence used to test each factor. Public
comparisons first evaluate the complete framework. Matched controls then
compare factor-wise evidence matching with a
shared read under two candidate evidence granularities. Targeted interventions
finally measure the effects of changing factor identity or factor-specific
operators. This order separates overall effectiveness, the source of the gain,
and the mechanism behind it.

\begin{table*}[!t]
  \centering
  \caption{Direct single-pass training-free ZS-CIR on FashionIQ validation.
  Average is the unweighted category mean. Bold and underlined values are the
  best and second best among the listed rows within each backbone block.}
  \label{tab:fashioniq_results}
  \setlength{\tabcolsep}{3.5pt}
  \renewcommand{\arraystretch}{1.05}
  \resizebox{\textwidth}{!}{%
  \begin{tabular}{@{}ll cc cc cc cc@{}}
    \toprule
    \multirow{2}{*}{\textbf{Backbone}} & \multirow{2}{*}{\textbf{Method}} &
    \multicolumn{2}{c}{\textbf{Shirt}} & \multicolumn{2}{c}{\textbf{Dress}} &
    \multicolumn{2}{c}{\textbf{Toptee}} & \multicolumn{2}{c}{\textbf{Average}} \\
    \cmidrule(lr){3-4}\cmidrule(lr){5-6}\cmidrule(lr){7-8}\cmidrule(lr){9-10}
    & & R@10 & R@50 & R@10 & R@50 & R@10 & R@50 & R@10 & R@50 \\
    \midrule
    \multirow{7}{*}{ViT-B}
      & CIReVL~\cite{karthik2024cirevl} (ICLR'24) & 28.36 & 47.84 & 25.29 & 46.36 & 31.21 & 53.85 & 28.29 & 49.35 \\
      & OSrCIR~\cite{tang2025osrcir} (CVPR'25) & 31.16 & 51.13 & 29.35 & 50.37 & 36.51 & 58.71 & 32.34 & 53.40 \\
      & CoTMR~\cite{sun2025cotmr} (ICCV'25) & 33.42 & 53.93 & 31.09 & 54.54 & 38.40 & 61.14 & 34.30 & 56.54 \\
      & CIReVL + SoFT~\cite{jung2026soft} (AAAIW'26) & 31.26 & 50.98 & 27.52 & 49.13 & 34.37 & 58.44 & 31.05 & 52.85 \\
      & SDR-CIR~\cite{sun2026sdrcir} (WWW'26) & \underline{36.41} & \underline{57.02} & \underline{36.84} & \underline{58.85} & \underline{43.14} & \underline{64.71} & \underline{38.80} & \underline{60.19} \\
      & STiTch~\cite{li2026stitch} (CVPR'26) & 25.22 & 44.16 & 18.59 & 40.16 & 25.97 & 47.61 & 23.26 & 43.98 \\
      \rowcolor{gray!15}
      & \textbf{TAPR (Ours)} & \textbf{39.45} & \textbf{59.22} & \textbf{38.87} & \textbf{61.43} & \textbf{45.95} & \textbf{67.36} & \textbf{41.42} & \textbf{62.67} \\
    \midrule
    \multirow{6}{*}{ViT-L}
      & CIReVL~\cite{karthik2024cirevl} (ICLR'24) & 29.49 & 47.40 & 24.79 & 44.76 & 31.36 & 53.65 & 28.55 & 48.57 \\
      & OSrCIR~\cite{tang2025osrcir} (CVPR'25) & 33.17 & 52.03 & 29.70 & 51.81 & 36.92 & 59.27 & 33.26 & 54.37 \\
      & CoTMR~\cite{sun2025cotmr} (ICCV'25) & 35.43 & 54.91 & 31.18 & 55.04 & 38.55 & 61.33 & 35.05 & 57.09 \\
      & CIReVL + SoFT~\cite{jung2026soft} (AAAIW'26) & 32.34 & 51.62 & 27.51 & 47.79 & 35.19 & 58.18 & 31.68 & 52.53 \\
      & SDR-CIR~\cite{sun2026sdrcir} (WWW'26) & \underline{41.02} & \underline{59.27} & \underline{37.04} & \underline{59.15} & \underline{44.47} & \underline{65.32} & \underline{40.84} & \underline{61.25} \\
      \rowcolor{gray!15}
      & \textbf{TAPR (Ours)} & \textbf{43.33} & \textbf{61.33} & \textbf{39.61} & \textbf{61.18} & \textbf{46.76} & \textbf{67.72} & \textbf{43.23} & \textbf{63.41} \\
    \midrule
    \multirow{6}{*}{ViT-G}
      & CIReVL~\cite{karthik2024cirevl} (ICLR'24) & 33.71 & 51.42 & 27.07 & 49.53 & 35.80 & 56.14 & 32.19 & 52.36 \\
      & OSrCIR~\cite{tang2025osrcir} (CVPR'25) & 38.65 & 54.71 & 33.02 & 54.78 & 41.04 & 61.83 & 37.57 & 57.11 \\
      & CoTMR~\cite{sun2025cotmr} (ICCV'25) & 38.32 & 62.24 & 34.51 & 57.36 & 41.90 & 64.30 & 38.25 & 61.32 \\
      & SDR-CIR~\cite{sun2026sdrcir} (WWW'26) & \underline{44.55} & \underline{62.37} & \underline{42.74} & \underline{63.41} & \underline{48.29} & \underline{69.71} & \underline{45.19} & \underline{65.16} \\
      & STiTch~\cite{li2026stitch} (CVPR'26) & 39.48 & 56.59 & 35.04 & 56.74 & 42.86 & 64.95 & 39.12 & 59.43 \\
      \rowcolor{gray!15}
      & \textbf{TAPR (Ours)} & \textbf{47.20} & \textbf{66.29} & \textbf{45.12} & \textbf{66.39} & \textbf{50.79} & \textbf{72.87} & \textbf{47.70} & \textbf{68.52} \\
    \bottomrule
  \end{tabular}%
  }
\end{table*}

\begin{figure*}[t]
  \centering
  \includegraphics[width=\textwidth]{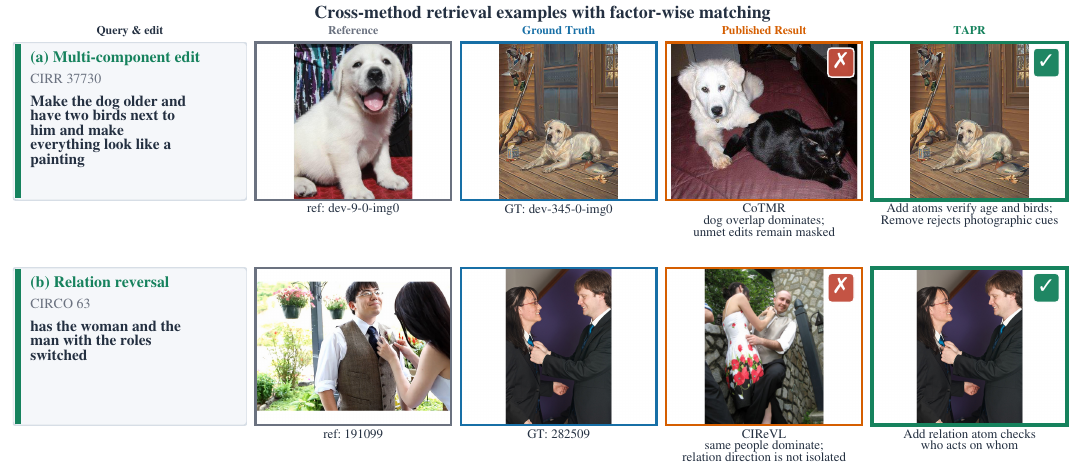}
  \caption{Cross-method retrieval examples on CIRR and CIRCO. For
  CIRR, CoTMR~\cite{sun2025cotmr} matches the dog but does not reject each unmet
  edit separately; \method{} verifies the Add atoms and penalizes Remove
  evidence for photographic content. For CIRCO, CIReVL~\cite{karthik2024cirevl}
  matches the people but not their relation direction; \method{} checks who acts
  on whom with a factor-specific probe. Red crosses and green checks mark incorrect and correct
  rank-1 results, respectively.}
  \label{fig:cross_method_rescue}
\end{figure*}

All \method{} results use frozen OpenCLIP image/text towers at ViT-B/32,
ViT-L/14, or
ViT-G/14 scale. Each image is cached as one global visual vector and a bank of
local visual vectors.
At 224 input resolution, ViT-B/32 keeps its native $7\!\times\!7$ grid
($M=49$); it is not expanded to 64 vectors. ViT-L/14 and ViT-G/14 use an
$8\!\times\!8$ pooled grid ($M=64$). Local interaction uses $\tau=0.02$, and a
frozen Qwen3.6-27B factorizer generates one factorized edit specification per query.
We restrict public comparison to methods that satisfy the training-free setting
in Sec.~\ref{sec:training_free_zscir} and directly score the gallery in a single
pass. Direct means that query-side records score the real gallery without
synthesizing a proxy target image. Single-pass means that all query records and
weights are fixed before gallery evaluation and each candidate receives its
final score in one gallery sweep. Fixed score fusion is allowed;
candidate-conditioned model calls, iterative retrieval, post-retrieval
verification, and rescue loops are excluded. All baseline numbers are taken
directly from the original papers. 

The selected methods cover distinct direct-scoring mechanisms. CIReVL captions
the reference, rewrites that caption under the edit, and retrieves with the
resulting target description~\cite{karthik2024cirevl}. OSrCIR instead uses
one-stage reflective multimodal reasoning to generate the target description
directly from the reference image and edit~\cite{tang2025osrcir}. LDRE + PDV-F starts from
multiple composed captions, transfers the prompt direction to both text and
reference-image embeddings, and fuses them~\cite{yang2024ldre,tursun2026pdv}.
CIReVL + SoFT
adds prescriptive and proscriptive global scores that reward required concepts
and penalize forbidden ones~\cite{jung2026soft}. STiTch refines a composed
caption through semantic transition and applies bidirectional transportation
for set-to-set matching~\cite{li2026stitch}. CoTMR derives global and
object-scale positive/negative descriptions through multimodal reasoning and
combines their frozen similarity scores~\cite{sun2025cotmr}. SDR-CIR uses
Selective CoT, anchors the query with reference-image features, and penalizes
the estimated reference-induced contribution~\cite{sun2026sdrcir}. \method{}
instead factorizes the query into Target, Add, Preserve, and Remove, encodes them
as one target description vector and factor-specific probe vectors, and uses
fine-grained multi-vector matching with distinct operators over one candidate
evidence bank containing global and local visual vectors.

Mechanism
analyses use OpenCLIP ViT-L/14. The controls in
Tables~\ref{tab:retrieval_factorial}--\ref{tab:role_operator} change only the
candidate evidence, factor assignment, or factor-to-operator dispatch stated in
each row.

\subsection{Overall Retrieval Effectiveness}
\label{sec:main_comparison}

Table~\ref{tab:main_results} reports CIRCO and CIRR test results, while
Table~\ref{tab:fashioniq_results} reports FashionIQ validation results. On
CIRCO, \method{} is highest at every cutoff and scale in the selected direct
single-pass comparison. At mAP@5 it improves over the best available comparator
by 23.0\%, 22.4\%, and 19.7\% for ViT-B/L/G. It also leads all three CIRR
full-gallery recalls and all three subset recalls at every backbone scale
within the selected comparison.

On FashionIQ, \method{} gives the highest category-specific and averaged
R@10 and R@50 at all three backbone scales, leading all 24 reported metrics.
The consistency across open-domain CIRCO/CIRR and category-specific FashionIQ
suggests that the benefit is not tied to one query style or visual domain. The
consistent gains from ViT-B to ViT-G further indicate that the factor-wise scorer
complements stronger frozen features rather than merely substituting for them.
Because every image and text tower remains frozen, this consistency cannot be
attributed to task-specific representation learning. Larger backbones change
the quality of the available evidence, yet the same asymmetric division among
Target, Add, Preserve, and Remove remains useful. This pattern is
consistent with pretrained features already encoding many relevant cues while
a holistic similarity score fails to use them according to their edit functions.

The metrics test complementary behaviors. CIRCO mAP measures consistent ranking
of multiple valid targets; CIRR full-gallery Recall@1 tests exact target
discrimination, and its subset metric focuses on related candidates. Gains on
all three show that \method{} neither relies on a restricted group nor only
benefits from multiple positives. FashionIQ further shows that the mechanism
transfers to attribute-centered edits across every category and cutoff: Target
maintains overall compatibility, while Add, Preserve, and Remove distinguish
specific edit requirements.

These comparisons establish overall effectiveness, but they cannot determine
whether the gains come from assigning distinct matching rules to different
edit factors or simply from representing each candidate with more vectors. The
matched analysis below separates these explanations.

Figure~\ref{fig:cross_method_rescue} explains why improving a global target
description does not ensure that every requested constraint controls ranking.
In panel (a), CoTMR's description-level matching is dominated by the dog and
scene, so a candidate can remain highly similar even when aging, the two birds,
and the painting style are missing. \method{} prevents this compensation: Add
atoms verify age and birds separately, while Remove atoms penalize photographic
evidence. In panel (b), CIReVL matches the correct people, but their shared
appearance can outweigh the reversed relation. \method{} represents the
requested actor--recipient direction as an independent Add atom, so who acts on
whom affects the score directly. The examples therefore illustrate the
mechanism behind the correction rather than only describing the retrieved
images.

\subsection{Evidence for Atomic Edit Factorization}
\label{sec:mechanism_analysis}

The remaining experiments test the central claim from mechanism to robustness.
We first separate factor-wise matching from candidate-evidence granularity,
then ablate edit factors and their operators, and finally vary the
local-interaction temperature and factor weights.

\subsubsection{Factor-Wise Matching Across Evidence Granularities}

Table~\ref{tab:retrieval_factorial} is the central matched control. All four
cells share the factorizer outputs and frozen weights. The
$\boldsymbol{\times}$ rows concatenate the nonempty A/P/R probes and apply one
unsigned shared read, removing both Preserve's reference-continuity term and
Remove's negative sign. The $\checkmark$ rows keep the same probes but apply
positive Add, reference-aware Preserve, and negative Remove operators. In the
Global Vector setting, reference continuity reduces to
global reference--candidate similarity.

\begin{table}[!t]
  \centering
  \caption{Effect of candidate-evidence granularity and factor-wise matching on
  ViT-L validation.}
  \label{tab:retrieval_factorial}
  \footnotesize
  \setlength{\tabcolsep}{3.0pt}
  \renewcommand{\arraystretch}{1.06}
  \resizebox{\columnwidth}{!}{%
  \begin{tabular}{@{}lc ccc@{}}
    \toprule
    \textbf{Candidate evidence} & \textbf{Factor-wise matching} &
    \shortstack{\textbf{CIRCO}\\\textbf{mAP@5}} &
    \shortstack{\textbf{CIRR}\\\textbf{Recall@1}} &
    \shortstack{\textbf{FashionIQ}\\\textbf{Macro R@10}} \\
    \midrule
    Global Vector & $\boldsymbol{\times}$ & 26.71 & 30.14 & 24.15 \\
    Global Vector & $\checkmark$ & 33.79 & 35.41 & 40.61 \\
    Global + Local & $\boldsymbol{\times}$ & 31.58 & 32.78 & 30.87 \\
    \rowcolor{gray!15}
    \textbf{Global + Local} & $\checkmark$
      & \textbf{37.17} & \textbf{38.28} & \textbf{43.23} \\
    \bottomrule
  \end{tabular}%
  }
\end{table}

Across both candidate-evidence settings, factor-wise matching improves the reported
metric on all three datasets. Using unrounded scores, the gains for the Global
Vector and Global + Local settings, respectively, are 7.08 and 5.58 points
on CIRCO, 5.26 and
5.50 on CIRR, and 16.46 and 12.36 on FashionIQ.
Adding local visual vectors under the shared read improves the three metrics by
4.88, 2.63, and 6.72 points; under factor-wise matching, it improves them by
3.38, 2.87, and 2.62 points. The interactions are $-1.50$, $+0.24$, and
$-4.10$, indicating partial overlap rather than additive benefits. The
factor-wise gain persists with one global visual vector, so it does not depend
on local candidate vectors. Local visual vectors nevertheless improve both
reads, and the strongest cell combines them with factor-wise matching. Thus,
Atomic Edit Factorization determines what evidence must be tested and how each
match affects ranking, while fine-grained multi-vector matching increases the
resolution of the evidence used for those tests.

The interactions show overlap rather than independent gains: factor-wise
matching can resolve some constraints globally, while local vectors make the
shared read less coarse. The matched ordering is decisive. Factor-wise matching
improves either evidence granularity, and Global + Local improves either
matching mode. The near-zero CIRR interaction further shows complementarity for
exact target discrimination.

\subsubsection{Edit Factors and Factor-Specific Operators}

Table~\ref{tab:role_components} removes one score from the T/A/P/R
combination, while Table~\ref{tab:role_operator} isolates factor assignment
and factor-specific matching operators.

\begin{figure*}[!t]
  \centering
  \includegraphics[width=\textwidth]{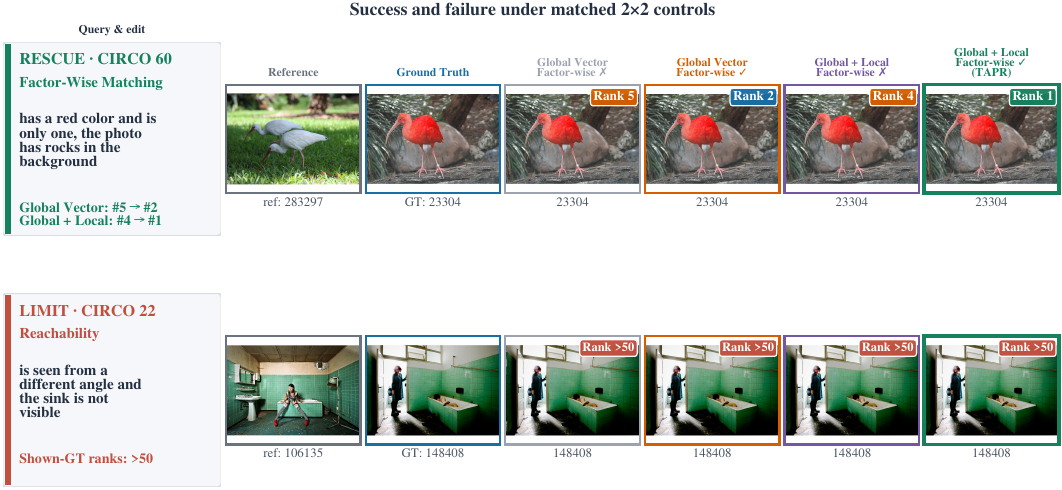}
  \caption{One success case and one failure case from the matched control on CIRCO validation.
  Each row fixes one ground-truth image across all four configuration columns.
  For CIRCO 60, image 23304 moves from ranks 5 to 2 and from 4 to 1. For CIRCO
  22, image 148408 remains outside the Top-50 in every cell.}
  \label{fig:success_failure}
\end{figure*}

\begin{table*}[!t]
\begin{minipage}[t]{0.48\textwidth}
  \vspace{0pt}
  \centering
  \caption{Ablation of the Target, Add, Preserve, and Remove factors on ViT-L
  validation.}
  \label{tab:role_components}
  \footnotesize
  \setlength{\tabcolsep}{3.0pt}
  \renewcommand{\arraystretch}{1.05}
  \begin{tabular*}{\linewidth}{@{\extracolsep{\fill}}lccc@{}}
    \toprule
    \textbf{Variant} &
    \shortstack{\textbf{CIRCO}\\\textbf{mAP@5}} &
    \shortstack{\textbf{CIRR}\\\textbf{Global R@1}} &
    \shortstack{\textbf{FashionIQ}\\\textbf{Macro R@10}} \\
    \midrule
    w/o Target & 22.02 & 24.64 & 28.89 \\
    w/o Add & 33.81 & 35.41 & 38.50 \\
    w/o Preserve & 35.76 & 35.89 & 34.77 \\
    w/o Remove & 35.83 & 37.08 & 42.07 \\
    \rowcolor{gray!15}
    \textbf{Full TAPR} & \textbf{37.17} & \textbf{38.28} & \textbf{43.23} \\
    \bottomrule
  \end{tabular*}
\end{minipage}\hfill
\begin{minipage}[t]{0.48\textwidth}
  \vspace{0pt}
  \centering
  \caption{Effects of factor assignment and factor-specific matching operators on
  ViT-L validation.}
  \label{tab:role_operator}
  \footnotesize
  \setlength{\tabcolsep}{3.0pt}
  \renewcommand{\arraystretch}{1.05}
  \begin{tabular*}{\linewidth}{@{\extracolsep{\fill}}lccc@{}}
    \toprule
    \textbf{Variant} &
    \shortstack{\textbf{CIRCO}\\\textbf{mAP@5}} &
    \shortstack{\textbf{CIRR}\\\textbf{Global R@1}} &
    \shortstack{\textbf{FashionIQ}\\\textbf{Macro R@10}} \\
    \midrule
    Factor reassignment (mean) & 31.05 & 32.25 & 37.09 \\
    w/o reference continuity & 35.60 & 36.12 & 34.77 \\
    Remove as reward ($+$) & 34.12 & 33.97 & 38.36 \\
    \rowcolor{gray!15}
    \textbf{Full TAPR} & \textbf{37.17} & \textbf{38.28} & \textbf{43.23} \\
    \bottomrule
  \end{tabular*}
\end{minipage}
\end{table*}

Target gives the largest contribution on all three datasets;
removing it lowers CIRCO mAP@5, CIRR Recall@1, and FashionIQ macro Recall@10 by
15.15, 13.64, and 14.34 points. Add is also consistently useful, while Preserve
matters most on FashionIQ. Remove is more metric dependent:
removing it lowers CIRR full-gallery Recall@1 by 1.20 points and lowers CIRR
subset Recall@1 from 71.06 to 69.29 and lowers the other two reported metrics.
The unequal effects reflect distinct jobs rather than interchangeable score
terms. Target supplies the holistic anchor, while Add, Preserve, and Remove test
what must appear, remain, or disappear. Preserve's larger effect
on FashionIQ is consistent with edits that retain garment identity, whereas
Remove helps when removed content distinguishes plausible distractors.
Smaller factor-specific corrections can therefore remain decisive among
candidates that share the same broad semantics.

Table~\ref{tab:role_operator} separates three properties that the composite
factor-wise switch in Table~\ref{tab:retrieval_factorial} changes together.
Factor reassignment permutes the same A/P/R probes among fixed operators while
preserving group sizes. Removing reference continuity deletes only the
$\tfrac12\mathcal L_\tau(V_\ell(I_r),V_\ell(I))$ term from Preserve and keeps
its half-weighted text term unchanged. Rewarding Remove flips only its sign;
the Preserve operator and all factor assignments remain intact. Each isolated
intervention lowers all three reported metrics. Thus, the intended factor
assignment, reference continuity, and negative Remove sign make distinct
contributions to these primary measures. The
reference-continuity effect is not uniform across CIRR measures: removing it
raises subset Recall@1 from 71.06 to 71.44 while lowering global Recall@1.
Factor reassignment is the most direct test of the factorization claim: it preserves
the available probes and matching operators but changes which meaning controls
which operator. Its consistent loss shows that extra descriptions alone are
insufficient; their semantic assignment matters. The continuity and sign
interventions explain why: preserved content needs a source-relative test, and
removed content must act as counter-evidence. Together,
Tables~\ref{tab:role_components} and~\ref{tab:role_operator} distinguish whether
a factor supplies useful evidence from whether that evidence is matched with the
operator its meaning requires. This explicit asymmetry is essential in a frozen
scorer that cannot learn it through an adapter.

Fig.~\ref{fig:success_failure} shows both the benefit and the boundary of this
design. In CIRCO 60, factor-wise matching improves the rank of ground-truth
image 23304 from 5 to 2 under Global Vector and from 4 to 1 under Global +
Local. The repeated improvement shows that factor-wise matching matters
independently of candidate evidence granularity. Local visual vectors further
move the result from rank 2 to rank 1, consistent with their function as finer
candidate evidence. In CIRCO 22, every configuration
misses all labeled
positives in the Top-50 for an edit involving viewpoint and an invisible sink.
Factor-wise scoring can organize available evidence, but it cannot recover a
target when the required cue is absent from the retrieved candidate set or is
not captured by the frozen visual representation.

\begin{figure*}[!t]
\begin{minipage}[t]{0.48\textwidth}
  \vspace{0pt}
  \centering
  \includegraphics[width=0.94\linewidth]{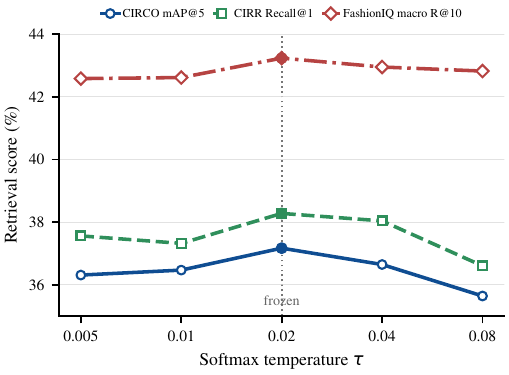}
  \caption{Softmax temperature sensitivity with OpenCLIP ViT-L/14. Only
  $\tau$ changes, and all three datasets use their complete validation sets.}
  \label{fig:temperature_sensitivity}
\end{minipage}\hfill
\begin{minipage}[t]{0.48\textwidth}
  \vspace{0pt}
  \centering
  \includegraphics[width=0.94\linewidth]{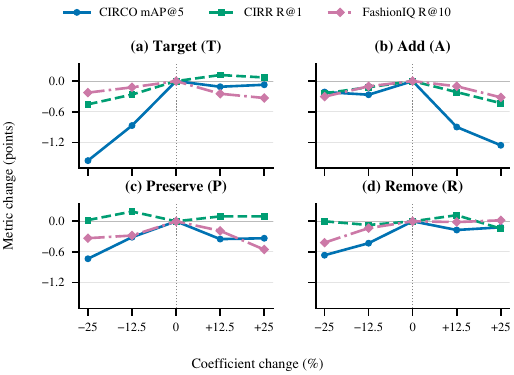}
  \caption{Target/Add/Preserve/Remove weight sensitivity with OpenCLIP
  ViT-L/14 on the complete CIRCO, CIRR, and FashionIQ validation sets.}
  \label{fig:weight_sensitivity}
\end{minipage}
\end{figure*}

\subsubsection{Softmax Temperature Sensitivity}

We vary only the local-interaction temperature $\tau$ over
$\{0.005,0.01,0.02,0.04,0.08\}$ while freezing the factorizer outputs, retrieval
weights, and all other settings. Fig.~\ref{fig:temperature_sensitivity} shows
that $\tau=0.02$ gives the highest score on all three reported metrics. Relative
to this point, the adjacent settings $0.01$ and $0.04$ are lower by 0.70
and 0.52 points on CIRCO, 0.96 and 0.24 on CIRR, and 0.62 and 0.28 on
FashionIQ. Temperature controls how strongly each probe concentrates on its
best-matching local visual vectors. A very small $\tau$ approaches hard local
selection and can overemphasize one incidental match, whereas a large $\tau$
spreads weight across weaker matches and dilutes factor-specific evidence. The
common optimum at $0.02$ therefore supports selective but non-degenerate local
aggregation. FashionIQ varies by only 0.65 points across the full grid, while
CIRCO and CIRR vary by 1.53 and 1.68 points, respectively, showing that the two
open-domain settings are more sensitive to local-evidence concentration in this
experiment.

\subsubsection{Factor-Weight Sensitivity}

We measure local sensitivity around the universal weights of the complete
factor-wise scorer. With OpenCLIP-L and $\tau=0.02$, we scale one frozen
coefficient at a time by $0.75$, $0.875$, $1.0$, $1.125$, or $1.25$ and apply
the same query-specific active-factor normalization.

Fig.~\ref{fig:weight_sensitivity} shows that, across the 16 non-anchor settings
for each dataset, the reported metric drops by at most 1.56 points on CIRCO, 0.45 on
CIRR, and 0.55 on FashionIQ. The best sampled neighbor differs from the frozen
point by $-0.07$, $+0.19$, and $+0.02$ points, respectively. The response is
therefore locally stable but not uniform: reducing Target by 25\% causes the
largest CIRCO drop, while increasing Preserve by 25\% causes the largest
FashionIQ drop.

These curves establish local rather than global robustness. CIRCO is most
sensitive to weakening the Target factor, whereas FashionIQ is most sensitive
to overemphasizing Preserve, consistent with the component
ablation. The conclusion therefore does not rely on one isolated weight setting.

Perturbation direction also matters: CIRCO degrades most when Target is
weakened or Add is overemphasized, whereas FashionIQ responds more to Preserve
than Remove near the anchor. These distinct profiles show that the four factors
are not interchangeable score terms. This directional evidence complements the
factor-removal results in Table~\ref{tab:role_components}.

\FloatBarrier

%% file: conclusion.tex
Training-free ZS-CIR requires inferring a target from a reference image and a
modification without task-specific training. We introduced \method{}, which
uses Atomic Edit Factorization to translate an entangled edit request into
structured visual evidence for retrieval. It separates the request into Target,
Add, Preserve, and Remove factors. The Target factor provides a holistic
anchor, while factor-specific probes capture content to be added, preserved,
or removed. Fine-grained multi-vector matching scores these probes against
global and local candidate representations, and factor-specific operators
integrate the resulting evidence according to each factor's function.
Experiments on CIRCO, CIRR, and FashionIQ with three frozen backbones
demonstrate consistent retrieval improvements. Controlled comparisons and
operator ablations further support the effectiveness of correct factor
assignment and factor-specific matching, while combining global and local
candidate representations consistently improves retrieval over global
representations alone. Remaining failures are mainly associated with
inaccurate factorization or insufficient representation of fine-grained
attributes and relations. Future work will improve edit-atom extraction and
region-aware evidence modeling while retaining the training-free retrieval
setting.